\documentclass[a4paper]{article}  

\usepackage{dx}
\usepackage{times}
\usepackage[T1]{fontenc}
\usepackage{ae,aecompl}
\usepackage{amsmath}
\usepackage{amssymb}
\usepackage{graphicx}
\usepackage{float}
\usepackage{subcaption}
\usepackage{epstopdf}
\usepackage[tableposition=top]{caption}
\usepackage{array}
\usepackage{url}
\usepackage[table]{xcolor}
\usepackage[procnames]{listings}

\title{Deploying Robots in Everyday Environments: Towards Dependable and Practical Robotic Systems}

\author%
{%
Alex Mitrevski, Santosh Thoduka, Argentina Ortega S{\'a}inz, Maximilian Sch{\"o}bel,\\
{\bf Patrick Nagel, Paul G. Pl{\"o}ger}, \and {\bf Erwin Prassler}\\
Hochschule Bonn-Rhein-Sieg, Sankt Augustin, Germany \\
e-mail: <aleksandar.mitrevski, santosh.thoduka, argentina.ortega,\\maximilian.schoebel, paul.ploeger, erwin.prassler>@h-brs.de \\
patrick.nagel@smail.inf.h-brs.de
}

\newcolumntype{M}[1]{>{\centering\arraybackslash}m{#1}}

\begin{document}

\maketitle
\thispagestyle{empty}
\pagestyle{empty}

\begin{abstract}
    Robot deployment in realistic dynamic environments is a challenging problem despite the fact that robots can be quite skilled at a large number of isolated tasks. One reason for this is that robots are rarely equipped with powerful introspection capabilities, which means that they cannot always deal with failures in a reasonable manner; in addition, manual diagnosis is often a tedious task that requires technicians to have a considerable set of robotics skills. In this paper, we discuss our ongoing efforts - in the context of the ROPOD\footnote{ROPOD is a Horizon 2020 project: \url{http://cordis.europa.eu/project/rcn/206247_en.html}} project - to address some of these problems. In particular, we (i) present our early efforts at developing a robotic black box and consider some factors that complicate its design, (ii) explain our component and system monitoring concept, and (iii) describe the necessity for remote monitoring and experimentation as well as our initial attempts at performing those. Our preliminary work opens a range of promising directions for making robots more usable and reliable in practice - not only in the context of ROPOD, but in a more general sense as well.
\end{abstract}

\section{Introduction}
\label{sec:introduction}

     Monitoring the behaviour of service robots and diagnosing failures in their components are essential to ensure the practical usefulness of robot deployment in everyday environments. This is particularly true in scenarios where the regular presence of a skilled technician is unlikely and where robots may come in close contact with human agents \cite{guiochet2017}.

\begin{figure}[tp]
        \centering
        \includegraphics[width=\linewidth]{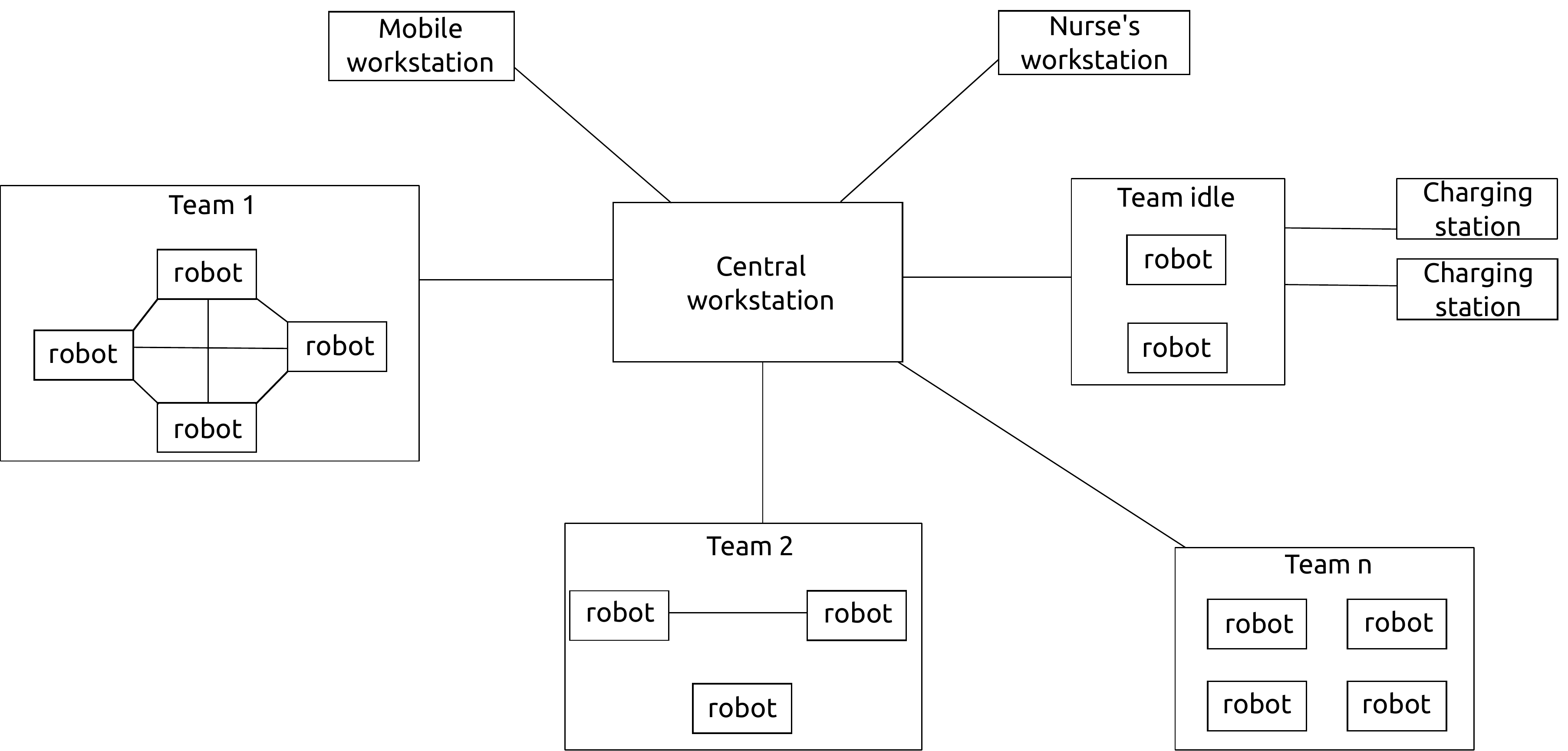}
        \caption{ROPOD system diagram}
        \label{fig:ropod_architecture}
    \end{figure}

    A scenario of this type is considered in the ROPOD project. The objective of ROPOD is to design and develop wheeled robots that can work together in order to transport various items in a hospital, such as beds, trolleys, and so forth. While performing these operations, the robots may periodically communicate with a central server and are also able to communicate amongst themselves. A high-level diagram of the ROPOD system is shown in Figure \ref{fig:ropod_architecture}.

    Transporting items in a hospital environment is challenging due to different reasons, such as the constant movement of people, the need of going in and out of elevators, and so forth. With that in mind, we focus on the fact that robots combine failure-prone mechanical, electronic, and software components, such that our objective is to define strategies for (i) \emph{online monitoring} of different state variables of these components for the purpose of failure detection, (ii) \emph{fault diagnosis} for analysing the effects of failures on the complete system, as well as (iii) \emph{remote operation monitoring} and \emph{remote experimentation}, which should aid human operators in the process of diagnosing system failures.

\begin{figure}[tp]
        \centering
        \includegraphics[width=\linewidth]{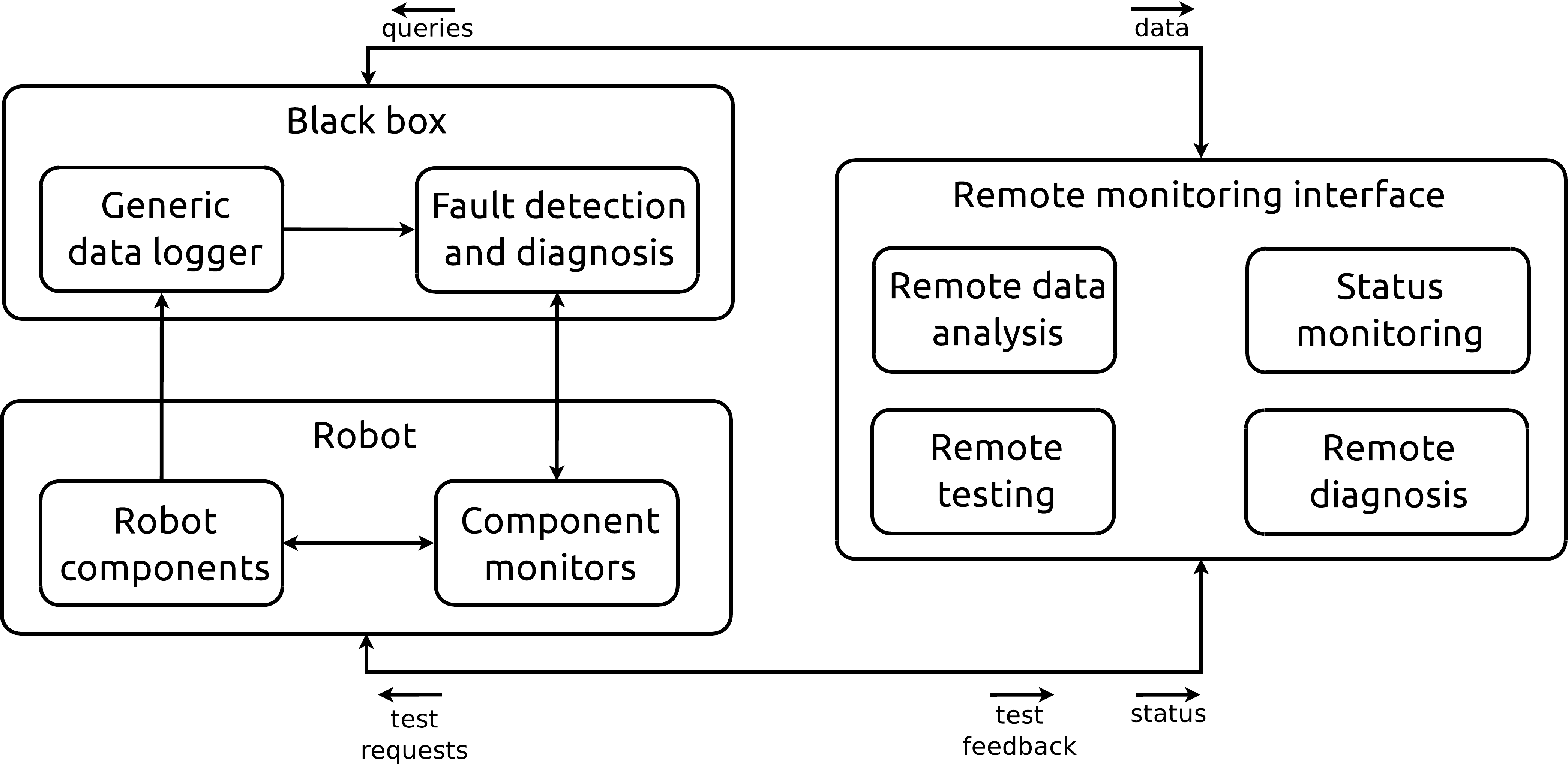}
        \caption{ROPOD monitoring and diagnosis concept}
        \label{fig:monitoring_and_diagnosis_concept}
    \end{figure}

    This paper describes our monitoring and diagnosis approach, which is illustrated in Figure \ref{fig:monitoring_and_diagnosis_concept}. In particular, we first develop a concept for a \emph{robotic black box}, a non-intrusive data logging device that can be added to robots and, in conjunction with state monitoring, can contribute to increasing their overall reliability. We further introduce a prototypical \emph{remote monitoring} interface that can be used for accessing black box data, robot status monitoring, and remote test execution. We finally focus on the problem of \emph{fault detection and diagnosis}, particularly its relation to the black box and the remote monitoring interface. It should be noted that this paper focuses on monitoring \emph{a single robot} as part of a robot team; we will extend this to multi-robot monitoring and diagnosis in our subsequent work.

\section{Related Work}
\label{sec:relatedWork}

\subsection{Data Logging}
\label{sec:relatedWork_logging}

    Logging data - both in robotics and in other domains - is generally not considered as an isolated problem, but logging is nevertheless an indispensable component in various contexts. In addition, several logging patents exist, particularly for flight data recorders and autonomous vehicles. We present a small subset of the literature here, particularly work related to robotics.

    Vasilescu et al. \cite{vasilescu2005} introduce a sensor architecture that can be used for underwater data collection. The main issue in this context is communication, both among sensor nodes deployed in different sea locations as well as between the nodes and a set of mobile robots that travel between locations. The main purpose of deploying sensor nodes underwater is data collection; a logging component is thus part of the setup, but data are logged at low frequencies (every 150s), so storage requirements are not an issue here.

    A black box architecture for discovering software failures is described by Elbaum and Munson \cite{elbaum2000}. This black box is composed of two components - a black box recorder, which logs transitions between system states and modules, and a black box decoder, which analyses the logged transitions and attempts to discover abnormalities in those. While the general idea presented here goes directly in line with our black box, one very distinguishing feature is that robots produce considerably more data, which contributes to a different set of requirements about storage and data lifetime.

    In the context of high-level planning and reasoning, Winkler et al. \cite{winkler2014} present the CRAM system, which integrates data logging into a full-fledged planning system. In conjunction with a logic-based query interface, the logged data allow monitoring the operation of a robot, which can subsequently be used for offline diagnosis of failures at the planning and execution level.

    In the above work, logging is generally done in a problem-specific, non-standard manner, which means that reusability is not of utmost importance. The objective of our work is to instead look at logging at a more abstract level in order to identify aspects that can be generalised regardless of the context; this is the main principle behind our black box design, which is presented in section \ref{sec:main_blackBox}.

\subsection{Monitoring, Fault Detection, and Fault Diagnosis}
\label{sec:relatedWork_monitoring}

    Data logging on its own is useful for analysing a system offline, but is not enough for discovering online abnormalities; a dependable system thus requires strategies for component/system monitoring, fault detection, and fault diagnosis as well. Here, we only discuss a few studies that are particularly important for our work; a more comprehensive review of fault detection and diagnosis can be found in \cite{pettersson2005} and \cite{khalastchi2018}.

    Visinsky et al. \cite{visinsky94} \cite{visinsky94_2} describe three fault diagnosis strategies, namely (i) Failure Mode, Effects, and Criticality Analysis (FMECA), (ii) fault trees, and (iii) expert systems. FMECA is a small extension of Failure Mode and Effect Analysis (FMEA), such that the idea is to list the ways in which components can fail along with the effects of those failures on the system and their expected frequency and severity. A fault tree organises events (for instance component failures) into a logical tree structure leading up to a top-level critical system failure; at runtime, the logical relationships can be used for finding out the likely events that have caused a certain failure. Expert systems, which typically consist of first-order logic rules, are particularly useful for runtime analysis, as one can make use of logic programming systems and their built-in inferencing algorithm for implementing a powerful diagnosis system.

    Dearden and Ernits \cite{dearden13} discuss the use of the discrete consistency-based Livingstone 2 system \cite{livingstone} for diagnosing faults in an underwater vehicle. The idea is to define a nominal model of the system based on a set of constraints on different variables describing the system's operation; faults can then be diagnosed by monitoring the variables at runtime and checking whether the constraints of the nominal model are satisfied.

    Probabilistic models, such as the banks of Kalman filters introduced by Roumeliotis et al. \cite{roumeliotis98}, can also be used for fault detection and diagnosis. In this case, a nominal model of a component or a system has to be developed, along with different fault models of the component or system in question. All of these models are represented by Kalman filters which are used for generating residuals based on the observed behaviour, such that faults can be diagnosed by comparing the residuals.

    Structural analysis as described by Blanke et al. \cite{blanke} is another technique that could be used for fault diagnosis. In structural analysis, a system is described by a set of constraints (which can have an algebraic, differential, or integral nature) and the variables involved in those; these are then connected by a so-called structural model/graph. By finding a graph matching on a given structural model, one can extract redundant relations that could then be used for calculating residuals and diagnosing faults.

    Golombek et al. \cite{golombek2010} treat fault detection as an anomaly detection problem, where anomalies are expected to manifest as changes in the signature of the data that flow through a system. Anomalies are detected by a method based on a probabilistic model of the time dependencies between various events that happen when a robot is performing specific tasks, which makes the detection method time- and task-dependent.

    A similar method in the context of multi-robot systems is presented in \cite{li2007} and \cite{li2009}. Just as \cite{golombek2010}, Li and Parker \cite{li2009} learn an event transition diagram that depends on a representation of the available data; furthermore, the detection of anomalies in both cases is based on a time-dependent analysis of states. The state transition diagram in \cite{li2009} compresses data from multiple information sources in a robot, such as sensor measurements and control commands. This represents the likelihood of transitions within a set of compressed versions of these features, where the compression is performed by data clustering.

    Our work builds on top of the related work in various different ways. More precisely, we are developing a diagnosis system similar to that described in \cite{visinsky94_2}. Component monitoring similar to that in \cite{dearden13} is also performed at different levels, i.e. not only at the component level of a single robot, but will also be done at the level of robots as components of multi-robot teams and the system as a whole. Finally, our general approach has various similarities with \cite{crestani2015}, such as the use of FMEA for identifying potential failures and using those to guide the detection and diagnosis.

\section{Data Logging, Fault Diagnosis, and Remote Monitoring}
\label{sec:main}

    As illustrated in Figure \ref{fig:monitoring_and_diagnosis_concept}, our dependability concept is based on three main aspects: (i) data logging using a robotic black box, (ii) component monitoring based on which fault detection and diagnosis are performed, and (iii) remote operation monitoring and experimentation.

\subsection{Robotic Black Box}
\label{sec:main_blackBox}

\begin{figure}[tp]
        \centering
        \includegraphics[width=\linewidth]{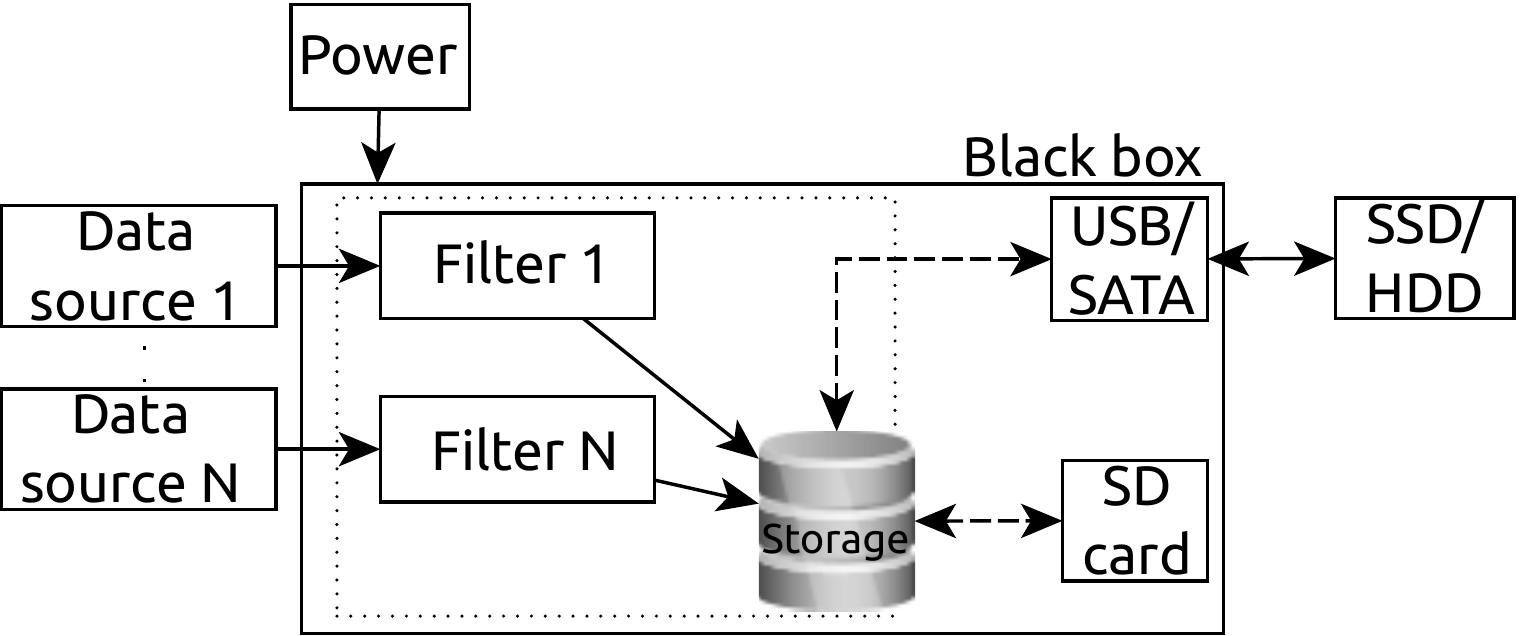}
        \caption{Generic black box concept}
        \label{fig:generic_black_box_concept}
    \end{figure}

    The robotic black box is supposed to operate as an independent embedded component that has the following properties: (i) the device has its own power supply and does not interfere with the operation of the system to which it is attached; (ii) the main functionalities of the black box are listening to data traffic from various sources, converting the data to a predefined format, and logging them in a manner that allows for easy access when diagnosing failures; and (iii) the device should be easily reconfigurable, such that it should be possible to use it on different robots with small modifications. The motivation for such a device comes from aircraft black boxes \cite{atsb2014}, which are indispensable when diagnosing aircraft failures and accidents, as well as from autonomous vehicles and the need of monitoring their operation at all times in order to avoid and, if necessary, explain fatal failures \cite{winfield2017}. The described robotic black box is supposed to be as general as possible; the concepts and part of the instrumentation are therefore robot-independent and are potentially standardisable. The general design of our robotic black box is illustrated in Figure \ref{fig:generic_black_box_concept}.

    Since different robots may have various communication interfaces (e.g. EtherCAT, WiFi, CAN bus), multiple sources that produce interface-specific data are likely to be available on a given robot. In other words, as there is no standard communication protocol in robotics, the black box must be designed with enough flexibility so that it can adapt to different interfaces on a given robot. In addition, each of the data sources might produce data that are likely to have different formats (e.g. ROS messages are different from ZMQ messages) and different frequencies, so the black box needs a means of understanding these data formats and converting them to a generic form. To address this problem, we associate each data source with a corresponding \emph{filter}, which (i) performs data filtering (e.g. ignores the data if they indicate an insignificant change in the signal), (ii) converts the data into a predefined format, and (iii) logs them on a data storage device.

    To make the black box as general as possible, we use a configuration file that specifies the available data sources and the data coming from those. The data filters use this configuration file in order to listen to the appropriate sources and log the data together with any available descriptive information about them, such as the variable names and their types. In principle, deploying the black box to a new robot would only involve creating filters for new data interfaces and adapting the configuration file to the data produced by the robot. A generic specification of a configuration file that listens to three different data sources - ROS messages, EtherCAT data, and Zyre messages - is shown below.
    \lstset{basicstyle=\ttfamily\scriptsize,
        backgroundcolor=\color{gray!10!white},
        frame=lines,
        breaklines=true,
        showstringspaces=false,
        caption={A sample black box configuration file}}
\begin{lstlisting}
- default_parameters:
    max_frequency: <int>
- ros:
    ros_master_uri: <string>
    topics:
        - topic:
            name: <string>
            type: <string>
            variable_names: <string>[]
            max_frequency: <int>
        - topic:
            name: <string>
            type: <string>
            variable_names: <string>[]
            max_frequency: <int>
- ethercat:
    interfaces:
        - interface_specification:
            interface: <string>
            input_message_size: <int>
            output_message_size: <int>
            number_of_slaves: <int>
            output_message_elements:
                - element:
                    name: <string>
                    type: <string>
                ...
            input_message_elements:
                - element:
                    name: <string>
                    type: <string>
                ...
            output_data_items:
                name: <string>
                variable_names: <string>[]
            input_data_items:
                name: <string>
                variable_names: <string>
- zyre:
    name: <string>
    groups: <string>[]
    message_types: <string>[]
\end{lstlisting}
    Another aspect that is not directly evident from the configuration file, but is hinted at by the lists of variable names and types, is that we log variables associated with a particular data item (e.g. a ROS odometry message) in an unrolled manner, namely data are always logged at a single level of abstraction even if the original item contains multiple levels; this requires item preprocessing at the time of logging, but makes data queries considerably simpler. In addition, we associate each logged item with a timestamp; this is generally the timestamp of the item itself, but we take a local timestamp if a certain item does not have its own timestamp. The message unrolling and the timestamp association are illustrated by the example below, which shows a ROS pose message and its format as stored on the black box.
    \lstset{basicstyle=\ttfamily\scriptsize,
        backgroundcolor=\color{gray!10!white},
        frame=lines,
        breaklines=true,
        showstringspaces=false,
        caption={A ROS pose message}}
\begin{lstlisting}
geometry_msgs/Point position
    float64 x
    float64 y
    float64 z
geometry_msgs/Quaternion orientation
    float64 x
    float64 y
    float64 z
    float64 w
\end{lstlisting}
    \lstset{basicstyle=\ttfamily\scriptsize,
        backgroundcolor=\color{gray!10!white},
        frame=lines,
        breaklines=true,
        showstringspaces=false,
        caption={Black box pose format}}
\begin{lstlisting}
double timestamp
double position_x
double position_y
double position_z
double orientation_x
double orientation_y
double orientation_z
double orientation_w
\end{lstlisting}

    Our current black box prototype for ROPOD\footnote{An implementation of our black box software can be found at \url{https://github.com/ropod-project/black-box}.} is a Raspberry Pi device and the data are logged in a MongoDB database that is stored on the SD card of the device; however, the actual nature of data storage may also be changed depending on the application requirements. For instance, it might be beneficial to use an SSD instead of an SD card for data storage if a significant data volume is expected (e.g. if camera images have to be recorded). This is particularly important when considering historical data, as the available storage space determines how often data need to be overwritten or copied offline. In the case of ROPOD, data will be copied to the central server once a robot finishes a mission and goes back to a charging station.

\subsection{Fault Detection and Diagnosis}
\label{sec:main_fdd}

    As we have seen in section \ref{sec:relatedWork}, fault diagnosis can be performed in a variety of ways, all of which have been successfully applied to various domains, both in robotics and outside of it. Our diagnosis concept makes use of this work; however, due to the complexity of multi-robot scenarios in which robots have to operate with little or no expert supervision, a single solution cannot work well in general. Due to this, different aspects of the existing models have to be combined in order to develop a more complete diagnosis system. In particular, our fault diagnosis concept is based on three complementary paradigms: component monitoring, system-level diagnosis, and remote monitoring.

    Our component-level monitoring approach is similar to the use of monitors in Livingstone 2 \cite{livingstone}. A component in the system could either be a hardware device or a software module; each such component has two levels of fault tolerance. The first is the  \emph{initialisation and configuration} level, during which the dependencies of each component are checked; in other words, a component fails to even initialise if the components on which it depends are not functional. By checking whether the dependencies of a given component are operational, we are able to pinpoint why the configuration fails, which is useful for performing diagnosis. The second level of fault tolerance is that of \emph{runtime monitoring}: just as in Livingstone 2, each component is associated with a normal operating state and a set of faulty states; transitions to the faulty states provide component-level diagnosis information.

    We model monitors as functions that get a certain input and produce a component status message as an output. Based on our abstraction, each component is associated with one or more monitors which may be redundant or may look at different aspects of the component; we refer to these monitors as component monitoring \emph{modes}. A configuration file for a given component, which has the general format shown below, thus specifies a list of modes and optionally a list of component dependencies:
    \lstset{basicstyle=\ttfamily\scriptsize,
        backgroundcolor=\color{gray!10!white},
        frame=lines,
        breaklines=true,
        showstringspaces=false,
        caption={Monitor configuration format}}
\begin{lstlisting}
name: <string> | required
modes: <string>[] | required
dependencies: <string>[] | optional
\end{lstlisting}
    In the monitor configuration file, \emph{modes} is a list of path names of component monitor mode configuration files. Each of these files defines the input-output mapping mentioned above and has the following format:
    \lstset{basicstyle=\ttfamily\scriptsize,
        backgroundcolor=\color{gray!10!white},
        frame=lines,
        breaklines=true,
        showstringspaces=false,
        caption={Monitor mode configuration format}}
\begin{lstlisting}
name: <string> | required
mappings:
    - mapping:
        inputs: <string>[] | required
        outputs: | required
            - output:
                name: <string> | required
                type: <string> | required
                expected: <bool> | <string> | <int> | <double> | optional
arguments: | optional
    - arg:
        name: <string> | required
        value: <bool> | <string> | <int> | <double> | required
\end{lstlisting}
    As can be seen, the mode configuration file also allows passing configuration arguments, such as thresholds that a monitor should use in order to determine whether a component is functioning properly; these arguments are however optional and may not be required by every monitor.

    The output produced by each component monitor is a string whose format is given as follows:
    \lstset{basicstyle=\ttfamily\scriptsize,
        backgroundcolor=\color{gray!10!white},
        frame=lines,
        breaklines=true,
        showstringspaces=false,
        caption={Status message format}}
\begin{lstlisting}
"robotId": "",
"monitorName": "",
"healthStatus":
{
    "output1_name": value,
    ...
    "outputN_name": value
}
\end{lstlisting}
    In this message, \emph{healthStatus} is a list of key-value pairs of the output names specified in the monitor configuration file along with the values corresponding to those.

    To illustrate the component monitoring configuration described above, we can consider an example in which a ROS-based robot has two laser scanners whose status we want to monitor. Let us suppose that we have two monitoring modes for the scanners, namely we can monitor whether (i) the hardware devices as such are recognised by the host operating system and (ii) the scanners are operational. A configuration file for this scenario would look as follows:
    \lstset{basicstyle=\ttfamily\scriptsize,
        backgroundcolor=\color{gray!10!white},
        frame=lines,
        breaklines=true,
        showstringspaces=false,
        caption={Laser monitor configuration}}
\begin{lstlisting}
name: laser_monitor
modes: [laser_monitors/device.yaml, laser_monitors/heartbeat.yaml]
dependencies: []
\end{lstlisting}
    Referring to the two monitor modes as device and heartbeat monitors, we will have the two monitor mode configuration files, which are shown below:
    \lstset{basicstyle=\ttfamily\scriptsize,
        backgroundcolor=\color{gray!10!white},
        frame=lines,
        breaklines=true,
        showstringspaces=false,
        caption={Laser device monitor mode configuration}}
\begin{lstlisting}
name: laser_device_monitor
mappings:
    - mapping:
        inputs: [/dev/front_laser]
        outputs:
            - output:
                name: front_laser_working
                type: bool
    - mapping:
        inputs: [/dev/rear_laser]
        outputs:
            - output:
                name: rear_laser_working
                type: bool
\end{lstlisting}
    \lstset{basicstyle=\ttfamily\scriptsize,
        backgroundcolor=\color{gray!10!white},
        frame=lines,
        breaklines=true,
        showstringspaces=false,
        caption={Laser heartbeat monitor mode configuration}}
\begin{lstlisting}
name: laser_heartbeat_monitor
mappings:
    - mapping:
        inputs: [/scan_front]
        outputs:
            - output:
                name: front_laser_working
                type: bool
    - mapping:
        inputs: [/scan_rear]
        outputs:
            - output:
                name: rear_laser_working
                type: bool
\end{lstlisting}
    As mentioned before, each monitoring mode is a function; in this particular example, the device monitor checks whether the paths specified by the inputs exist on the file system, while the heartbeat monitor checks whether ROS messages are published by the laser scanner drivers on the topics specified by the input.

    While component-level monitors provide direct information about the operation of individual components, unmodelled or incorrectly modelled states of operation may lead to faults being propagated to other components; in such cases, system-level diagnosis should help in discovering the causes of faults. As a starting point for system-level diagnosis, it is informative to perform an FMEA of the system. In ROPOD, we perform FMEA on two different levels, namely (i) by observing the system as a whole, which includes the central server, the robots, an interface that the hospital personnel will use to interact with the system, as well as all the software that ties everything together and (ii) by observing each robot as a system on its own, which is made up of various other hardware and software components. Based on the FMEA and the failure classes defined in \cite{steinbauer2013}, \cite{wienke2017}, and \cite{carlson2004}, rules and models for rule-based diagnosis and structural analysis respectively can be created. We already have an early prototype of a Prolog-based expert system that associates symptoms to a set of known effects on the components and on the ROPOD system as a whole. Simple examples of rules encoded in the system are given below:
    \begin{align*}
        broken(X) :- \; &robot(Y), initialisation\_errors(Y),\\
        &wheel(Z), freely\_rotating(Z),\\
        &motor(X).
    \end{align*}
    \begin{align*}
        short\_circuit(X) :- \; &robot(X),\\
        &driver\_initialising(X),\\
        &initialisation\_shutdown(X).
    \end{align*}
    Such a diagnosis system could be used for both offline and online diagnosis. For instance, the system could generate fault hypotheses that should be explored more closely, which is particularly important in the case of reduced component observability or incompletely specified monitoring models. Our early diagnosis prototype does not yet include a structural analysis component, but we aim to include that after having a more complete FMEA.

    Remote monitoring is the third aspect on which our diagnosis concept is based, but this is described separately in the following section.

\subsection{Remote Monitoring and Testing}
\label{sec:main_remoteMonitoring}

    As mentioned in the introduction, the objective in the ROPOD project is to deploy robots in a hospital environment where a system technician will not be available all the time. Because of that, faulty robots will often have to be diagnosed by a remote technician, who will need to analyse the robots without having direct access to them. The presence of a robotic black box on each individual robot in principle aids this process, but an interface is required for remotely accessing these data.

\begin{figure}[t]
        \centering
        \includegraphics[scale=0.17]{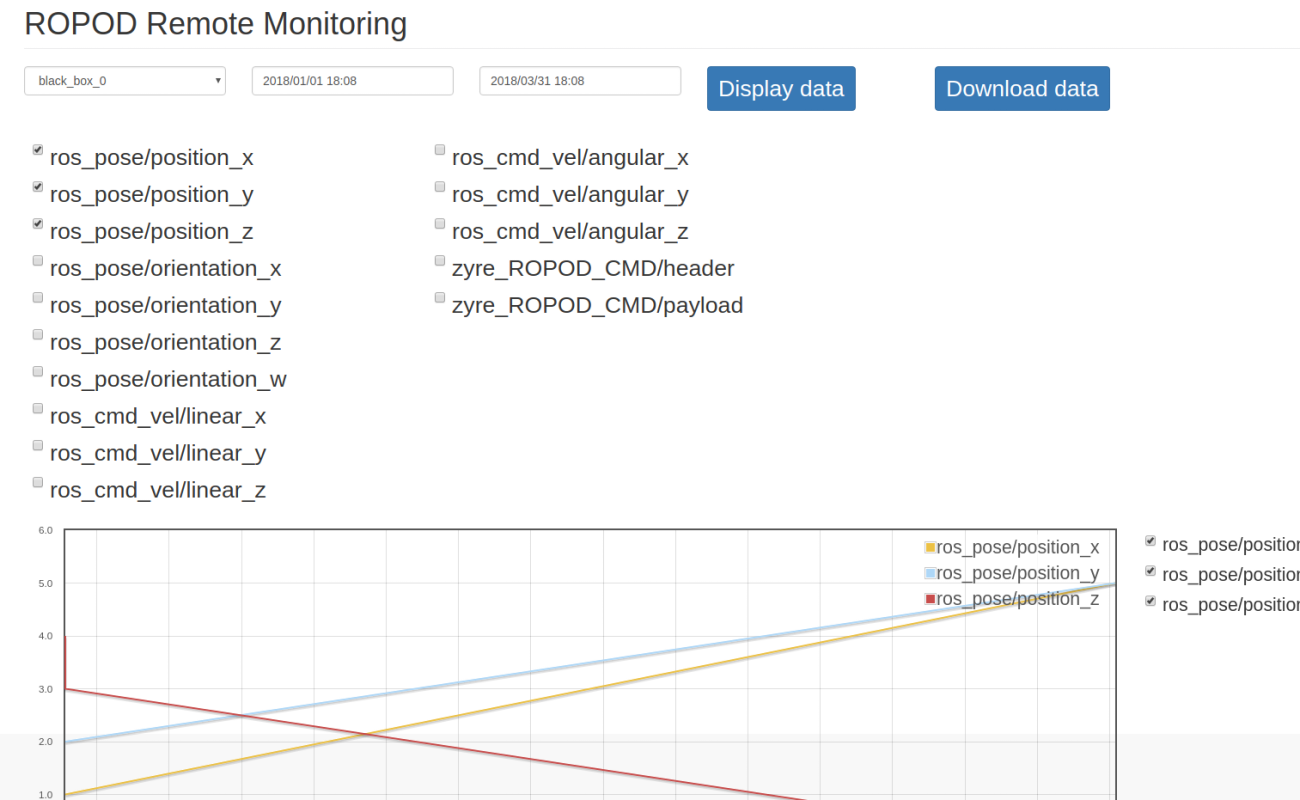}
        \caption{Remote data analysis: Black box data for a given time period can be visualised and optionally downloaded}
        \label{fig:remote_data_analysis}
    \end{figure}

    In order to allow both skilled and non-skilled technicians to analyse the operation of robots, we are developing a remote monitoring interface that offers diverse functionalities for diagnosis and testing. First of all, we can visualise black box data from robots that are connected to the hospital network (not all areas of the hospital will have network coverage, so we do not expect all robots to be online all the time). Not all data might be suitable for being visualised however, so we additionally allow bulk data download for offline analysis. This aspect of our remote monitoring interface is shown in Figure \ref{fig:remote_data_analysis}.

\begin{figure}[t]
        \centering
        \includegraphics[scale=0.17]{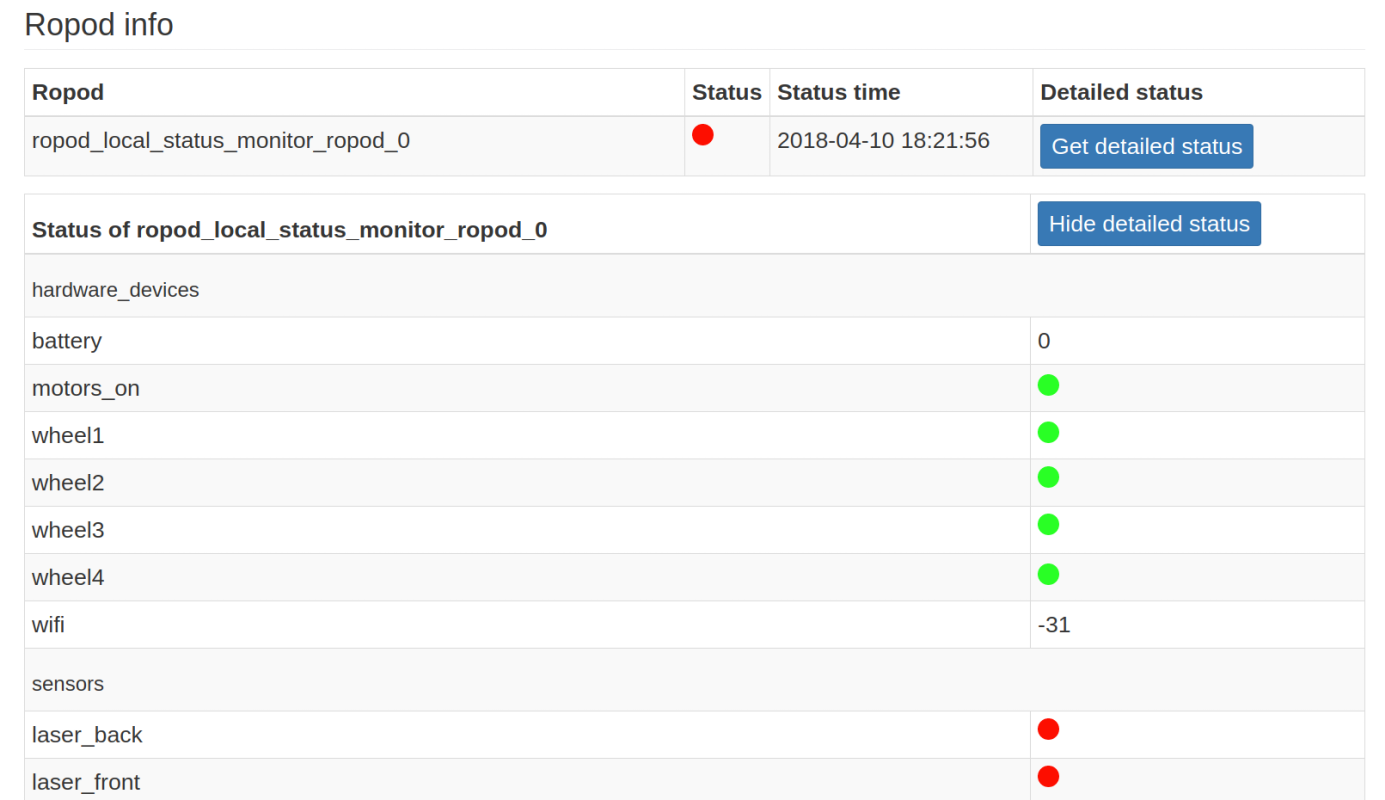}
        \caption{Status monitoring: The component monitors running on each robot continuously advertise the status of each component}
        \label{fig:status_monitoring}
    \end{figure}

    While visualising data does shed light about what might be wrong with a robot, interpreting the data may not be an easy task, particularly for less-skilled remote technicians. In addition, selecting which data to visualise is not trivial if there is little prior information about the failures that need to be diagnosed. For this reason, we additionally visualise the status of the component monitors (shown in Figure \ref{fig:status_monitoring}), which is expected to be advertised continuously; this provides direct information about which components are operational and which ones are not behaving as expected.%

    Analysing data about the actual operation of robots is necessary for both online and offline operation monitoring, but this method of remote monitoring cannot be directly used for preemptively eliminating failures caused by human factors (e.g. new software component that causes unforeseen effects on the system). Software tests and simulations are useful in this context, but since robot systems have to deal with hardware as well, these testing methods may not always provide full information about the effects a change would have on the system. Because of this, we have also integrated functionalities for performing remote experiments and unit tests into our remote monitoring interface in a manner similar to \cite{pitzer2012}. Such tests provide various opportunities, such as testing failure hypotheses and verifying that new functionalities do not break any existing ones.

\begin{figure}[t]
        \centering
        \includegraphics[width=\linewidth]{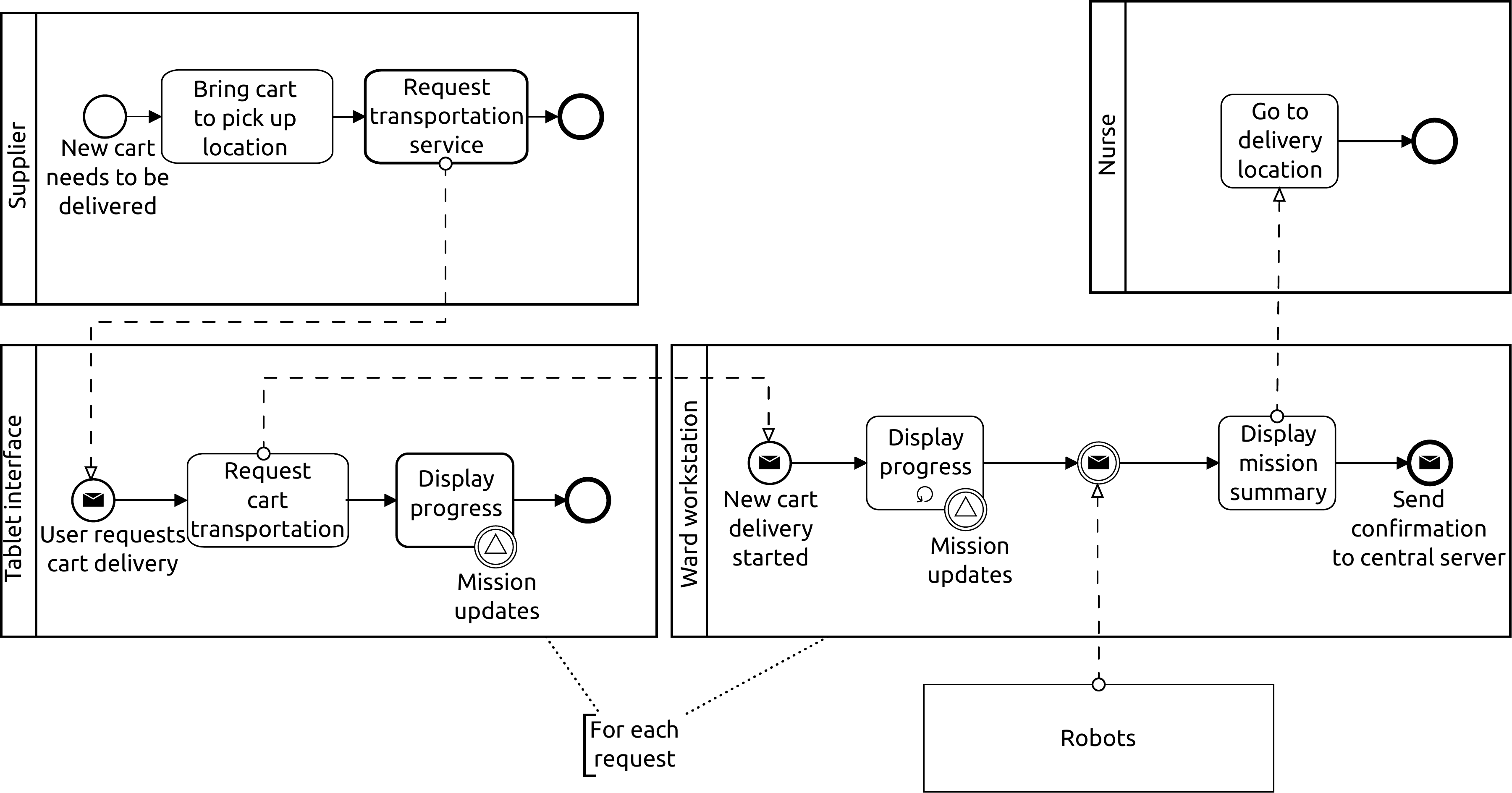}
        \caption{Autonomous cart transportation BPMN workflow diagram}
        \label{fig:mobidik_bpmn}
    \end{figure}

	When performing standardised tests, experimental reproducibility needs to be guaranteed for the tests to be of any value. For that purpose, test models should explicitly encode all aspects about the experimental setup and workflow, thus ensuring that the same experimental procedure will be followed during a test regardless of the experimenters. This is particularly important when performing remote experiments that involve on-site operators who may not be skilled robot users. We address this by creating experimental workflow diagrams using the Business Process Model and Notation (BPMN), although UML activity diagrams could be used as well; in particular, each of the tests that we can run remotely has an associated diagram or a set of diagrams, which an on-site operator can use for noticing any deviations from the expected test behaviour. A sample diagram for the case of transporting an autonomous cart in a hospital is shown in Figure \ref{fig:mobidik_bpmn}. This diagram makes all aspects about the scenario explicit; for instance, it can be seen that a cart should be delivered to a designated pickup location before robots are requested to deliver it and similarly, a nurse should wait for the cart at the pickup location.

\section{Discussion and Conclusions}
\label{sec:discussionAndConclusions}

    This paper has discussed our ongoing work for increasing the dependability of robotic systems that are deployed to environments where robots need to work alongside human agents and are not continuously supervised by expert operators. The dependability concept presented here is based on three main principles: (i) data logging with the help of a robotic black box, (ii) component/system monitoring and fault diagnosis, and (iii) remote operation monitoring and experimentation.

    In the context of ROPOD, future work includes performing long-running tests in the target hospital environment, which we need in order to examine the storage requirements of the black box more closely and get better acquainted with remote diagnosis. While initial testing was done under lab experiments that mimic the hospital scenario in general and the autonomous cart transportation use case in particular, this does not entirely represent the realistic deployment scenario.

    More generally, we would like to apply our monitoring and diagnosis concept to the other robots we work with so that we explore its generality more closely. Some ongoing efforts in this direction have already started in the context of our RoboCup@Home team, particularly in the sense of high-level component monitoring, but we are working on extending this to monitoring low-level components as well. We would additionally like to deploy our robotic black box to the robot used in a project dealing with telepresence robots in remote, difficult-to-reach locations. Finally, we are currently extending our remote monitoring interface so that we can (i) interact with the rule-based diagnosis engine through it and (ii) obtain better visualisation of the state of the robots.

\section{Acknowledgements}
\label{sec:acknowledgements}

    ROPOD is an Innovation Action funded by the European Commission under grant no. 731848 within the Horizon 2020 framework program. We would also like to thank Mohammadali Varfan, who has contributed to the remote monitoring interface presented in this paper.

\bibliographystyle{unsrt}
\bibliography{references}

\end{document}